%% file: main.tex
\documentclass{article}
\usepackage{graphicx} %
\usepackage{geometry}

\usepackage{fullpage}
\usepackage{pgfplots}
\pgfplotsset{compat=1.15}
\usetikzlibrary{calc, arrows,intersections, patterns}

\usepackage[
ruled,
linesnumbered,
]{algorithm2e}

\usepackage{macro}
\newcommand{\vertleq}[2]{ \underset{ \overset{ \rotatebox{-90}{$\,<$} }{\displaystyle{#2}} }{\displaystyle{#1}} }

\newcommand{\act}{\mathrm{act}}

\title{A Fast, Robust Elliptical Slice Sampling Implementation \\ for Linearly Truncated Multivariate Normal Distributions}
\author{
    Kaiwen Wu
    \and
    Jacob R. Gardner\thanks{
        Both authors are with the Department of Computer and Information Science, University of Pennsylvania.
        Correspondence to: \texttt{kaiwenwu@seas.upenn.edu}, \texttt{jacobrg@seas.upenn.edu}.
    }
}
\date{}

\begin{document}

\maketitle

\begin{abstract}
Elliptical slice sampling, when adapted to linearly truncated multivariate normal distributions, is a rejection-free Markov chain Monte Carlo method.
At its core, it requires analytically constructing an ellipse-polytope intersection.
The main novelty of this paper is an algorithm that computes this intersection in $\mathcal{O}(m \log m)$ time, where $m$ is the number of linear inequality constraints representing the polytope.
We show that an implementation based on this algorithm enhances numerical stability, speeds up running time, and is easy to parallelize for launching multiple Markov chains.
\end{abstract}

\section{Introduction}
\input{paper/introduction}

\section{Elliptical Slice Sampling for Truncated Normal Sampling}

\begin{figure}
\centering
\begin{minipage}[t]{0.32\linewidth}
\vspace{0pt}
    \input{figures/illustration/ellipse.tex}
\end{minipage}
\hfill
\begin{minipage}[t]{0.66\linewidth}
\vspace{0pt}
\begin{minipage}[c]{\linewidth}
    \input{figures/illustration/legend}
\end{minipage}
\caption{
An ellipse $\xv_t \cos\theta + \nuv_t \sin\theta$ whose angle $\theta \in [0, 2\pi]$ increases counterclockwise.
The next iterate $\xv_{t+1}$ is sampled from the ellipse-polytope intersection, as shown in red.
The intersection consists of two disjoint elliptical arcs.
The left one is represented by $[\frac78\pi, \frac98\pi]$ and
the right one is represented by $[0, \frac18\pi] \cup [\frac74\pi, 2\pi]$.
}
\label{fig:ellipse}
\end{minipage}
\end{figure}
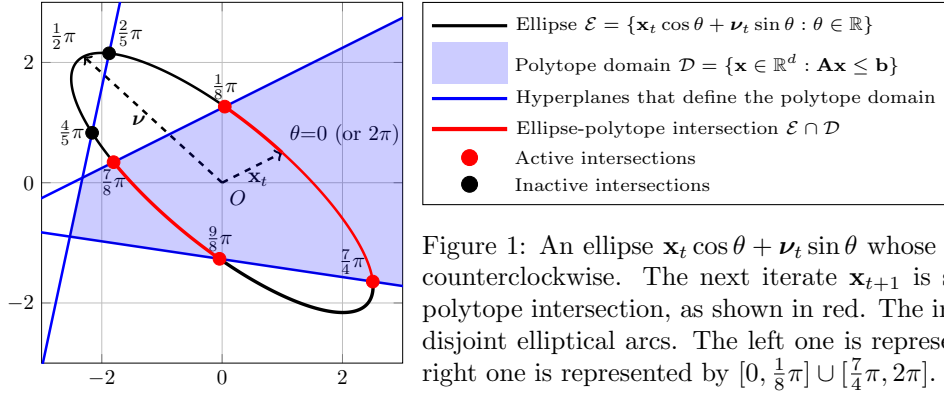

\input{algorithms/ess}

Linear elliptical slice sampling \parencite[\eg,][]{fagan2016elliptical,gessner2020integrals} is a specialized elliptical slice sampling method \parencite{murray2010elliptical} for linearly truncated multivariate normal distributions by exploiting the structure of the polytope domain.
In the $t$-th iteration, we sample a multivariate normal random variable $\nuv_t \sim \Nc(\zero, \Iv)$ and form an ellipse
\begin{equation}
\label{eq:ellipse}
    \Ec = \cbb{\xv_t \cos\theta + \nuv_t \sin\theta: \theta \in \Rb}.
\end{equation}
The next iterate $\xv_{t+1}$ is sampled from the ellipse-polytope intersection $\Ec \cap \Dc$, \ie, the parts of the ellipse that lie inside the polytope domain.
This intersection can be constructed analytically by exploiting the polytope structure, and thus no rejection sampling is needed.
See \Cref{fig:ellipse} for an illustration and \Cref{alg:ess} for the pseudocode.
The stationary distribution of this Markov chain is exactly the truncated normal distribution \parencite{murray2010elliptical}.
The arising questions are, of course, how to ``analytically construct'' the ellipse-polytope intersection $\Ec \cap \Dc$ and how to do it efficiently.

Note that the polytope domain itself is the intersection of $m$ halfspaces:
\begin{equation*}
    \Dc = \bigcap_{i=1}^{m} \Hc_i = \bigcap_{i=1}^{m} \cbb{\xv \in \Rb^d: \av_i^\top \xv \leq b_i},
\end{equation*}
where $\av_i$ is the $i$-th row of $\Av$.
Thus, the ellipse-polytope intersection $\Ec \cap \Dc$ reduces to constructing each ellipse-halfspace intersection $\Ec \cap \Hc_i$, which does admit an analytical construction.

The intersection of the ellipse and the $i$-th halfspace $\Hc_i$ is an elliptical arc.
The end points of the elliptical arc are identified by the intersection angles, \ie, the roots of the trigonometry equation
\begin{equation}
\label{eq:trigonometry-equation}
    \av_i^\top \xv_t \cos\theta + \av_i^\top \nuv_t \sin\theta = b_i,
\end{equation}
which typically indeed has two distinct roots $\alpha_i$ and $\beta_i$ in closed-forms (see \S\ref{sec:trigonometry}).
It is possible that \eqref{eq:trigonometry-equation} has no root or only a single root, but we defer these edge cases to \S\ref{sec:edge-cases}.
Without loss of generality, we assume all intersection angles $\alpha_i$ and $\beta_i$ are converted into $[0, 2\pi]$ by, if necessary, adding or subtracting a multiple of $2 \pi$.
In addition, we assume $\alpha_i$ is strictly smaller than $\beta_i$.
A simple observation is that the ellipse-halfspace intersection $\Ec \cap \Hc_i$ is precisely represented by the union of two disjoint intervals:
\begin{align*}
    I_i = \sbb{0, \alpha_i} \cup \sbb{\beta_i, 2\pi}.
\end{align*}
Even though the ellipse-halfspace intersection $\Ec \cap \Hc_i$ is evidently a continuous curve, its interval representation $I_i$ has two disjoint segments due to periodicity.
The point $\xv_t$ is represented by two distinct angles $0$ and $2\pi$.
If we ``glue'' together $0$ and $2\pi$, then $I_i$ can be viewed as a ``continuous'' interval.\footnote{
Converting all intersection angles to a different range, \eg, $[-\pi, \pi]$, does not necessarily makes all interval representations continuous and does make the downstream algorithmic problem any easier.
}
Intersecting all $I_i$'s gives the interval representation of the ellipse-polytope intersection:
\[
    I_\act = \bigcap_{i=1}^{m} I_i,
\]
which we call the active intervals.
Note that the plural form is used because $I_\act$ may consist of several disjoint intervals, each of which is an active interval.
There is an one-to-one correspondence between angles in the active intervals (except the repetition of $\theta = 0$ and $\theta = 2\pi$) and points in the ellipse-polytope intersection.

\subsection{The Obvious Algorithm?}

\input{paper/naive}

\subsection{Likelihood Testing}
\input{paper/likelihood-testing}

\section{A Simple Method for the Ellipse-Polytope Intersection}

\input{paper/method}

\section{Experiments}
\label{sec:experiments}
\input{paper/experiments}

\section{Related Work}
\input{paper/related}

\section{Conclusion}
\input{paper/conclusion}

\section*{Acknowledgement}
KW expresses his gratitude to Yuxin Sun, who came up with the lower bound argument in \S\ref{sec:general-case}.
KW also thanks Geoff Pleiss for introducing this problem, which makes this work possible.

\printbibliography

\clearpage

\appendix
\input{appendix}

\end{document}

%% file: paper/introduction.tex
Let $\xv \sim \Nc(\zero, \Iv)$ be a $d$-dimensional standard normal random variable.
This paper is concerned with sampling from the truncated multivariate normal distribution
\begin{align*}
p(\xv) =
\begin{cases}
\frac{1}{Z} \phi(\xv) & \xv \in \Dc,
\\
0 & \xv \notin \Dc,
\end{cases}
\end{align*}
where $\phi(\xv) \propto \exp\bb[\big]{-\frac12 \xv^\top \xv}$ is the standard normal density, $Z = \int_{\xv \in \Dc} \phi(\xv) \diff \xv$ is a normalization constant, and the domain $\Dc = \cbb{\xv \in \Rb^d: \Av \xv \leq \bv}$ is a polytope defined by $m$ linear inequalities with $\Av \in \Rb^{m \times d}$ and $\bv \in \Rb^m$.
We assume the polytope domain has a non-empty interior but is not necessarily bounded.
The standard normal assumption is without loss of generality, since non-standard normal distributions can be handled by a change of variable, as shown in \S\ref{sec:nonstandard-normal}.

Truncated normal sampling has numerous applications in machine learning and statistics, with recent ones in skew Gaussian processes \parencite[\eg,][]{benavoli2021unified} and preferential Bayesian optimization \parencite{takeno2023towards,benavoli2021unified}.
In addition, truncated normal sampling is a key building block of sophisticated numerical methods estimating integrals related to truncated normal distributions \parencite{gessner2020integrals}.

This paper, in particular, focuses on elliptical slice sampling \parencite{murray2010elliptical} for truncated normal sampling.
Each iteration of elliptical slice sampling computes the intersection of an ellipse and the polytope domain, from which the next iterate is sampled.
Since this ellipse-polytope intersection can be constructed analytically, the proposal is always accepted and no rejection sampling is needed \parencite{fagan2016elliptical,gessner2020integrals}.
In principle, this sampling method is tuning-free and particularly suitable for high dimensional truncated normal distributions.

However, the devil is in the details.
Analytically constructing the ellipse-polytope intersection is easier said than done, despite its conceptual simplicity.
We will show that all existing implementations share a worst-case time complexity of $\Oc(m^2)$, which scales poorly as the number of constraints increases.
Moreover, existing implementations have complex control flows, which makes it hard, if not impossible, to parallelize on GPUs.
Indeed, to the best of our knowledge, there is no batch implementation of elliptical slice sampling to this date, which is likely due to the programming complexity.

\textbf{Contributions.}
We develop a new algorithm computing the ellipse-polytope intersection that has a better time complexity and is easier to implement.
The algorithm runs in $\Oc(m \log m)$ time faster than the existing implementations.
Moreover, this algorithm has a simple control flow and is particularly amenable to GPU parallelism.
As a result, we are able to parallelize thousands of independent Markov chains easily.
We also discuss how to handle edge cases and numerical instability.
Experiments show that our implementation accelerates truncated normal sampling massively in high dimensions.

%% file: figures/illustration/ellipse.tex
\begin{tikzpicture}[scale=0.8]
\begin{axis}[
    x=1cm,y=1cm, grid=major,
    xmin=-3, xmax=3,
    ymin=-3, ymax=3,
]

\def\xtx{1}
\def\xty{0.5}
\def\nux{-2.3}
\def\nuy{2.1}

\def\ex#1{{\xtx * cos(deg(#1)) + \nux * sin(deg(#1)}}
\def\ey#1{{\xty * cos(deg(#1)) + \nuy * sin(deg(#1)}}

\def\point#1{\ex{#1}, \ey{#1}}

\coordinate[label=below right:$O$] (O) at (0, 0);
\coordinate[label=above right:$\theta{=}0$ (or $2\pi$)] (xt) at (\xtx, \xty);
\coordinate[label=above left:$\frac12\pi$] (nu) at (\nux, \nuy);

\draw[->,very thick, dashed] (O) -- node[below, pos=0.6] {$\xv_t$} (xt);
\draw[->,very thick, dashed] (O) -- node[below, pos=0.6] {$\nuv$} (nu);

\newcommand{\pointA}{\point{pi * 0.125}}
\newcommand{\pointB}{\point{pi * 0.875}}
\coordinate[label=above:$\frac18\pi$](A) at (\pointA);
\coordinate[label=below:$\frac78\pi$] (B) at (\pointB);

\newcommand{\pointC}{\point{pi * 1.125}}
\newcommand{\pointD}{\point{pi * 1.650}}
\coordinate[label=above:$\frac98\pi$] (C) at (\pointC);
\coordinate[label=above left:$\frac74\pi$] (D) at (\pointD);

\newcommand{\pointE}{\point{pi * 0.400}}
\newcommand{\pointF}{\point{pi * 0.800}}
\coordinate[label=above right:$\frac25\pi$] (E) at (\pointE);
\coordinate[label=left:$\frac45\pi$] (F) at (\pointF);

\draw [name path=L1,very thick, blue] ($(B)!6cm!(A)$) -- ($(A)!6cm!(B)$);
\draw [name path=L2,very thick, blue] ($(C)!6cm!(D)$) -- ($(D)!6cm!(C)$);
\draw [name path=L3,very thick, blue] ($(F)!6cm!(E)$) -- ($(E)!6cm!(F)$);

\coordinate (southwest) at (-3, -3);
\coordinate (southeast) at (3, -3);
\coordinate (northeast) at (3, 3);
\coordinate (northwest) at (-3, 3);

\path[name path=bottom] (southwest) -- (southeast);
\path[name path=top] (northwest) -- (northeast);
\path[name path=right] (northeast) -- (southeast);

\path[name intersections={of=L1 and right, by={i1}}];
\path[name intersections={of=L1 and L3, by={i2}}];

\path[name intersections={of=L2 and L3, by={i3}}];
\path[name intersections={of=L2 and right, by={i4}}];

\draw[fill=blue, opacity=0.2] (i1) -- (i2) -- (i3) -- (i4);

\addplot[domain=pi * 0.000:pi * 0.125,samples=100,red,  ultra thick] (\ex{x}, \ey{x});
\addplot[domain=pi * 0.125:pi * 0.875,samples=100,black,very thick] (\ex{x}, \ey{x});
\addplot[domain=pi * 0.875:pi * 1.125,samples=100,red,  ultra thick] (\ex{x}, \ey{x});
\addplot[domain=pi * 1.125:pi * 1.650,samples=100,black,ultra thick] (\ex{x}, \ey{x});
\addplot[domain=pi * 1.650:pi * 2.000,samples=100,red,  very thick] (\ex{x}, {\ey{x}});

\draw[fill,red] (\pointA) circle (3pt);
\draw[fill,red] (\pointB) circle (3pt);
\draw[fill,red] (\pointC) circle (3pt);
\draw[fill,red] (\pointD) circle (3pt);
\draw[fill,black] (\pointE) circle (3pt);
\draw[fill,black] (\pointF) circle (3pt);
\end{axis}

\end{tikzpicture}

%% file: figures/illustration/legend.tex
\begin{tikzpicture}
\scriptsize
\matrix[draw,below left] {
  \draw[black,very thick] (0, 0) -- (1, 0) node[right,black] {Ellipse $\Ec = \{\xv_t \cos\theta + \nuv_t \sin\theta: \theta \in \Rb\}$}; \\
  \fill[blue, opacity=0.2] (0, 0) rectangle (1, -0.55) node[above right,black,opacity=1] {Polytope domain $\Dc = \cbb{\xv \in \Rb^d: \Av \xv \leq \bv}$}; \\
  \draw[blue, very thick] (0, 0) -- (1, 0) node[right, black] {Hyperplanes that define the polytope domain}; \\
  \draw[red,ultra thick] (0, 0) -- (1, 0) node[right,black] {Ellipse-polytope intersection $\Ec \cap \Dc$}; \\
  \draw[fill,red] (0.5, 0) circle (3pt); \node[right] at (1, 0) {Active intersections}; \\
  \draw[fill,black] (0.5, 0) circle (3pt); \node[right] at (1, 0) {Inactive intersections}; \\
};

\end{tikzpicture}

%% file: algorithms/ess.tex
\begin{algorithm}[b]
\DontPrintSemicolon

Initialize $\xv_0 \in \Dc$\;
\For{$t = 1, 2, \cdots$}{
    sample $\nuv_t \sim \Nc(\zero, \Iv)$ and form an ellipse $\Ec = \cbb{\xv_t \cos\theta + \nuv_t \sin\theta: \theta \in \Rb}$\;
    compute the active intervals $I_\act \subseteq [0, 2\pi]$ corresponding to the ellipse-polytope intersection $\Ec \cap \Dc$\;
    sample uniformly $\theta \sim I_\act$\;
    $\xv_{t+1} = \xv_t \cos\theta + \nuv_t \sin\theta$\;
}
\caption{Elliptical Slice Sampling for Linearly Truncated Multivariate Normal Distributions}
\label{alg:ess}
\end{algorithm}

%% file: paper/naive.tex
\input{algorithms/naive}
Some readers might think computing the active intervals is trivial, given that \Cref{alg:naive} obviously solves the problem according to the definition of $I_\act$ and appears to run in linear time.

In general, the intermediate variable $I_\act^{(i)}$ in \Cref{alg:naive} consists of several disjoint segments---it could be the union of several disjoint intervals.
Thus, implementing \Cref{line:one-step-intersection} needs to enumerate all segments, which necessarily adds an inner for loop to the algorithm.
In the worst case, the intermediate variable $I_\act^{(i-1)}$ could have as many as $i$ segments, and thus the time complexity could be as bad as $\Omega(m^2)$.
We direct readers to \S\ref{sec:naive-running-time} where we construct a worst-case input and show that \Cref{alg:naive} indeed takes $\Omega(m^2)$ operations.

Besides its quadratic time complexity, \Cref{alg:naive} has several additional drawbacks.
First, implementing the interval intersection according to its definition, even though elementary, is extremely tedious.
Second, \Cref{alg:naive} has to process each interval $I_i$ sequentially, which is not friendly to GPU implementation.
Third, it is hard to implement a batch version to run multiple Markov chains in parallel, since the intermediate variable $I_\act^{(i)}$ may have different number of segments across the batch.

%% file: algorithms/naive.tex
\begin{algorithm}[t]
\DontPrintSemicolon
\KwIn{$I_i = \sbb{0, \alpha_i} \cup \sbb{\beta_i, 2\pi}$ for $i = 1, 2, \cdots, m$}
\KwOut{$I_\act = \cap_{i=1}^{m} I_i$}
$I_\act^{(0)} = \sbb{0, 2\pi}$\;
\For{$i = 1, 2, \cdots, m$}{
    $I_\act^{(i)} = I_\act^{(i-1)} \cap \bb[\big]{\sbb{0, \alpha_i} \cup \sbb{\beta_i, 2\pi}}$\;
    \label{line:one-step-intersection}
}
\Return $I_\act^{(i)}$\;
\caption{Constructing the Active Intervals by Brute Force}
\label{alg:naive}
\end{algorithm}

%% file: paper/likelihood-testing.tex
The hard part of computing the active intervals is identifying those active intersection angles, since some of the ellipse-halfspace intersections do not contribute to the active intervals.
In \Cref{fig:ellipse}, for example, $\theta = \frac25 \pi$ and $\theta = \frac45 \pi$ are not active, and thus they have no effect on the ellipse-polytope intersection.
Constructing the active intervals is easy once the active intersection angles are given.

\textcite{gessner2020integrals} identify the active angles by likelihood testing.
Assume the $2 m$ intersection angles are distinct and sorted in ascending order:
\[
    0 < \theta_1 < \theta_2 < \cdots < \theta_m < \cdots < \theta_{2m} < 2\pi.
\]
Define the likelihood function
\[
\ell(\theta) =
\begin{cases}
1 & \Av\bb{\xv_t \cos\theta + \nuv_t \sin\theta} \leq \bv, \\
0 & \text{otherwise}.
\end{cases}
\]
The likelihood function outputs one if and only if the point represented by $\theta$ is inside the domain.
Then, for every intersection angle $\theta_i$, \textcite{gessner2020integrals} detect the likelihood jump
\(
    \Delta_i = \ell(\theta_i + \epsilon) - \ell(\theta_i - \epsilon)
\),
where $\epsilon > 0$ is a small positive scalar.
There are two possible outcomes:
(a) $\theta_i$ is inactive and should be discarded if $\Delta_i = 0$;
(b) $\theta_i$ is an active angle if $\Delta_i = \pm 1$.
Evaluating the likelihood function requires two matrix-vector multiplications, which costs $\Oc(md)$ time, and testing $2m$ angles takes a total of $\Oc(m^2d)$ time.
Though, this complexity can be improved to $\Oc(m^2)$ if using cached values of $\Av \xv_t$ and $\Av \nuv_t$, which are computed anyway when solving the trigonometry equation \eqref{eq:trigonometry-equation}.

The constant $\epsilon$ has to be tuned carefully.
The likelihood function, when evaluated numerically, may produce false outputs due to floating-point errors if $\epsilon$ is too small.
On the flip side, $\epsilon$ should be no larger than the minimum gap $\min_{i \neq j} \abs{\theta_i - \theta_j}$.
An improved version in BoTorch \parencite{balandat2020botorch} automates the choice of the constant $\epsilon$ by testing the mid point $\frac12 \bb{\theta_i + \theta_{i+1}}$.
An angle $\theta_i$ is active if and only if the two adjacent mid points $\frac12(\theta_{i-1} + \theta_i)$ and $\frac12(\theta_i + \theta_{i+1})$ have different likelihoods.
However, likelihood testing fundamentally requires that the intersection angles have to be distinct.

%% file: paper/method.tex
\input{algorithms/best}

We now present our $\Oc(m \log m)$ algorithm computing the active intervals, which has a one-to-one correspondence with the ellipse-polytope intersection $\Ec \cap \Dc$.
For presentation simplicity, we temporarily assume the equation \eqref{eq:trigonometry-equation} has two distinct roots $\alpha_i < \beta_i$ in $[0, 2\pi]$ for all $1 \leq i \leq m$.
We will deal with the annoying edge cases when this assumption is violated in \S\ref{sec:edge-cases}.
As we will see soon, it almost takes no extra effort computationally to handle the edge cases.

\subsection{General Case}
\label{sec:general-case}

Sorting $\alpha_i$'s in ascending order yields
\begin{equation}
\label{eq:sorted_left_endpoint}
0 \leq \vertleq{\alpha_{i_1}}{\beta_{i_1}} \leq \vertleq{\alpha_{i_2}}{\beta_{i_2}} \leq \cdots \leq \vertleq{\alpha_{i_k}}{\beta_{i_k}} \leq \cdots \leq \vertleq{\alpha_{i_m}}{\beta_{i_m}} \leq 2\pi.
\end{equation}
We emphasize that $\beta_{i_k}$ is not necessarily monotonic in $k$.
Our algorithm is based on the observation below.
\begin{restatable}[]{proposition}{ClosedFormIntersection}
\label{thm:closed-form-intersection}
Let $\alpha_i < \beta_i$ for all $i \in [m]$ and
let $\cbb{\alpha_{i_k}}_{k=1}^{m}$ be sorted in ascending order as in \eqref{eq:sorted_left_endpoint}.
Then, the active intervals $I_\act = \cap_{i=1}^{m} \bb[\big]{[0, \alpha_i] \cup [\beta_i, 2\pi]}$
have an equivalent representation
\begin{align*}
I_\act = [0, \alpha_{i_1}] \cup \bb*{\bigcup_{k=2}^{m} [\gamma_{k-1}, \alpha_{i_k}]} \cup [\gamma_m, 2\pi],
\end{align*}
where $\gamma_k = \max\cbb{\beta_{i_1}, \beta_{i_2}, \cdots, \beta_{i_k}}$ is the cumulative max until $\beta_{i_k}$.
We interpret the interval $[\gamma_{k-1}, \alpha_{i_k}]$ as an empty set if $\gamma_{k-1} > \alpha_{i_k}$.
\end{restatable}
\begin{proof}[Proof Sketch]
The sorted angles $\cbb{\alpha_{i_k}}_{k=1}^{m}$ divide $[0, 2\pi]$ into $m + 1$ disjoint segments:
\begin{align*}
\sbb{0, 2\pi} =
[0, \alpha_{i_1}] \cup
(\alpha_{i_1}, \alpha_{i_2}] \cup
\cdots \cup
(\alpha_{i_{m-1}}, \alpha_{i_m}] \cup
(\alpha_{i_m}, 2\pi].
\end{align*}
Note the trivial identity $I_\act = I_\act \cap [0, 2\pi]$ because $I_\act$ itself is a subset of $[0, 2\pi]$.
Then, $I_\act$ is constructed by computing the intersection of $I_\act$ with each segment:
\begin{align*}
I_\act
= I_\act \cap \sbb{0, 2\pi}
=
\underbrace{
    \bb[\big]{I_\act \cap \sbb{0, \alpha_{i_1}}}
}_{\text{part one}}
\cup
\underbrace{
    \bb*{\bigcup_{k=1}^{m-1} \bb[\big]{I_\act \cap (\alpha_{i_k}, \alpha_{i_{k + 1}}]}}
}_{\text{part two}}
\cup
\underbrace{
    \bb[\big]{I_\act \cap \sbb{\alpha_{i_m}, 2\pi}}
}_{\text{part three}},
\end{align*}
where each part turns out to have a closed-form expression.
Part one is shown equal to $[0, \alpha_{i_1}]$,
part three is shown equal to $[\gamma_m, 2\pi]$,
and each segment in part two has the identity
\begin{align*}
I_\act \cap \sbb{\alpha_{i_k}, \alpha_{i_{k+1}}} = [\gamma_{k-1}, \alpha_{i_k}], \quad k = 2, 3, \cdots, m.
\end{align*}
We direct readers to \S\ref{sec:proofs} for a complete proof.
\end{proof}

\Cref{thm:closed-form-intersection} gives a closed-form expression for the active intervals $I_\act$, which yields \Cref{alg:best}.
Despite a somewhat lengthy proof, the idea and the final expression are both extremely simple.
The algorithm is more numerically robust comparing to likelihood testing, because it only relies on pairwise comparisons, not floating-point operations like addition and multiplication.
Thus, it does not introduce any floating-point errors.
For example, \Cref{alg:best} is guaranteed to work even when some angles happen to be identical.
The time complexity is $\Oc(m \log m)$, faster than the bruce force method and likelihood testing.
In addition, it is simple to program, amenable to GPU parallelism, and easy to batch, since the sorting and cumulative max operations are well supported in every popular machine learning package nowadays.

\textbf{A Complexity Lower Bound?}
The $\Oc(m \log m)$ complexity is likely to be optimal, due to the existence of a reduction from sorting to constructing the active intervals.
Consider sorting $m$ distinct numbers $c_1, c_2, \cdots, c_m$ in $(0, 2\pi)$.
Let $\epsilon$ be a positive number that is strictly smaller than $\min_{i \neq j}\abs{c_i - c_j}$.
Then, consider intersecting the following intervals
\[
    I_i = [0, c_i] \cup [c_i + \epsilon, 2\pi], \quad i = 1, 2, \cdots, m.
\]
By \Cref{thm:closed-form-intersection}, the intersection is of the form $[0, c_{i_1}] \cup \bb[\big]{\cup_{k=2}^{m} [c_{i_{k-1}} + \epsilon, c_{i_k}]} \cup [c_{i_m} + \epsilon, 2\pi]$.
In particular, the interval $[c_{i_{k-1}} + \epsilon, c_{i_k}]$ ``reveals'' the predecessor of $c_{i_k}$.
Therefore, any algorithm computing the active intervals immediately results in a sorting algorithm.
Thus, computing the active intervals is at least as ``hard'' as sorting, which has a well-known $\Oc(m \log m)$ complexity lower bound.
Note that this reduction does not give a rigorous proof, because the $\Oc(m \log m)$ complexity lower bound only holds for comparison-based sorting.
Nonetheless, this argument hints that it is unlikely that a faster algorithm exists.

\subsection{Edge Cases: No Intersection and a Single Intersection}
\label{sec:edge-cases}

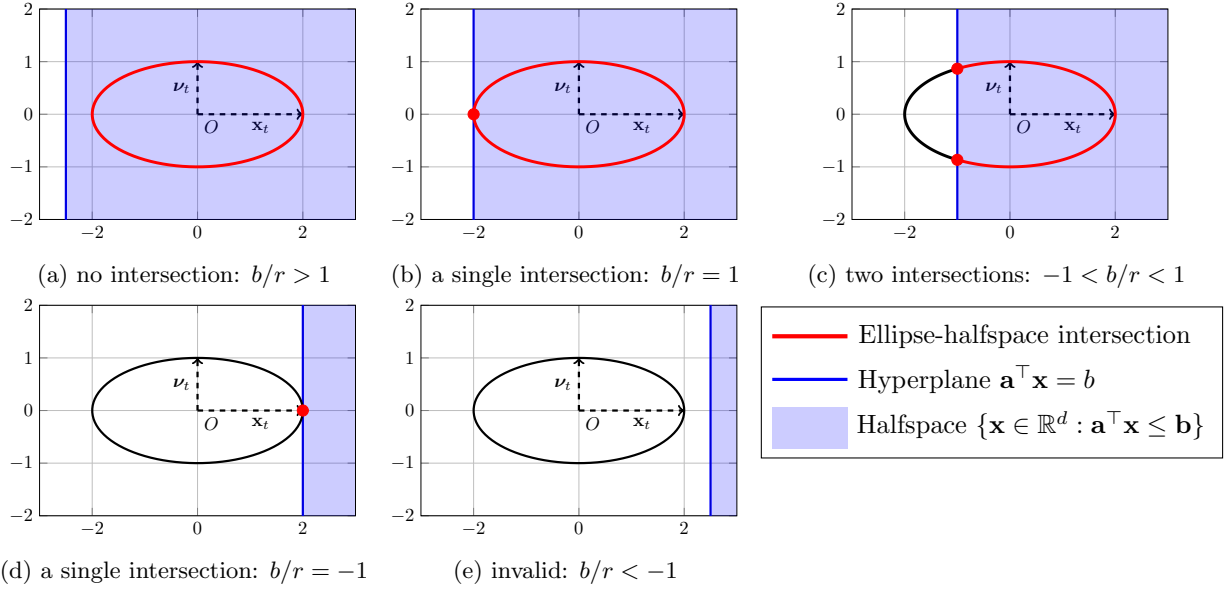
\begin{figure}
\centering
\begin{subfigure}[t]{0.3\linewidth}
\centering
\input{figures/intersections/ratio_gt_1}
\caption{no intersection: $b / r > 1$}
\label{fig:illustration-no-intersection}
\end{subfigure}
\begin{subfigure}[t]{0.3\linewidth}
\centering
\input{figures/intersections/ratio_=_1}
\caption{a single intersection: $b / r = 1$}
\label{fig:illustration-single-intersection-easy}
\end{subfigure}
\begin{subfigure}[t]{0.38\linewidth}
\centering
\input{figures/intersections/ratio_lt_1}
\caption{two intersections: $-1 < b / r < 1$}
\end{subfigure}

\begin{subfigure}[t]{0.3\linewidth}
\centering
\vspace{0pt}
\input{figures/intersections/ratio_=_-1}
\caption{a single intersection: $b / r = -1$}
\label{fig:illustration-single-intersection-hard}
\end{subfigure}
\begin{subfigure}[t]{0.3\linewidth}
\centering
\vspace{0pt}
\input{figures/intersections/ratio_lt_-1}
\caption{invalid: $b / r < -1$}
\end{subfigure}
\begin{minipage}[t]{0.38\linewidth}
\centering
\vspace{.5em}
\input{figures/intersections/legend}
\end{minipage}
\caption{
The number of intersections between the ellipse $\xv \cos\theta + \nuv \sin\theta$ and a hyperplane $\av^\top \xv = b$ depends on the ratio $b / r$, where $r = \sqrt{(\av^\top \xv_t)^2 + (\av^\top \nuv_t)^2}$.}
\label{fig:illustration-number-of-intersections}
\end{figure}

This section discusses how to handle the edge cases when some trigonometry equation \eqref{eq:trigonometry-equation} has either zero or a single root.
The number of roots is determined by the ratio
\begin{equation*}
    \frac{b}{r} \in [-1, +\infty),\; \text{where}\; r = \sqrt{(\av^\top \xv_t)^2 + (\av^\top \nuv_t)^2}.
\end{equation*}
We direct readers to \S\ref{sec:trigonometry} for an explanation why the ratio $b / r$ is no smaller than $-1$.
There are three outcomes based on the ratio $b / r$: no root if $b / r > 1$, a single root if $b / r = \pm 1$, two distinct roots if $-1 < b / r < 1$.
Each case is visualized in \Cref{fig:illustration-number-of-intersections}.

The nonexistence of the root, when $b / r > 1$, implies that the ellipse has no intersection with the hyperplane, which happens only when the entire ellipse is contained in the halfspace represented by the hyperplane, as shown in \Cref{fig:illustration-no-intersection}.
The corresponding linear inequality constraint has no effect on the ellipse-polytope intersection and thus can be simply ignored.
Alternatively, we artificially set the intersection angles $\alpha_i = \beta_i = 0$.
It is easy to verify that \Cref{alg:best} still works properly.
Adding these ``padding'' angles makes sure \Cref{alg:best} always receives the same number of intersection angles, making batching easier.

A single root happens when $b / r = \pm 1$.
In this case, the hyperplane is tangent to the ellipse, as shown in \Cref{fig:illustration-single-intersection-easy,fig:illustration-single-intersection-hard},
which in principle does not happen in practice.
With floating-point arithmetic, the division $b / r$ could almost never be exactly $\pm 1$.
We do not (explicitly) handle this edge case and pretend nothing bad would happen.

A more interesting question to think about is what would happen if $b / r$ is approximately $\pm 1$, \ie, the hyperplane is ``almost'' tangent to the ellipse.
To answer this question, we have to differentiate $b / r \approx 1$ and $b / r \approx -1$.
The easy case is $b / r \approx 1$, where the entire ellipse is ``almost'' contained by the corresponding halfspace (depending on $b / r$ approaches $1$ from the left or the right).
As a result, the corresponding linear inequality constraint can be safely ignored (just like what we did when $b / r > 1$).

The hard one is $b / r \approx -1$, when the ellipse is almost outside the polytope domain.
This edge case is likely to cause numerical issues, because the ellipse-polytope intersection is a \emph{tiny} elliptical arc.
We are forced to sample $\xv_{t+1}$ from a tiny segment on the ellipse, \eg, \Cref{fig:illustration-single-intersection-hard}.
A small floating-point error may shoot $\xv_{t+1}$ outside the domain.
We defer the issue to the next section, where we discuss numerical stability collectively.

\subsection{Numerical Stability}
\label{sec:numerical-stability}
Linear elliptical slice sampling is unstable when $\xv_t$ is too close to the domain boundary.
A small floating-point error in the angular domain $[0, 2\pi]$ may be amplified when mapping the angle $\theta$ to a high dimensional vector $\xv_t \cos\theta + \nuv_t \sin\theta$.
Thus, the Markov chain may arrive at an infeasible point in some extreme situations due to floating-point errors, which leads to two major consequences.

First, an infeasible point violating the constraints can never be a sample from the truncated distribution, and thus has to be discarded.
Second, \Cref{alg:best} does not produce the ``correct'' active intervals when $\xv_t$ is infeasible because its assumption is violated.
It is possible to modify \Cref{alg:best} to handle the cases when $\xv_t$ lies outside the domain.
However, modifications like this are ad hoc, ugly, and makes the implementation unnecessarily complicated, which completely contradicts with our original goal of finding a simple algorithm.

We present two tricks that enhance the numerical stability.
First, we trim the active intervals by a small constant, replacing every interval of the form $[l, u]$ with $[l + \epsilon, u - \epsilon]$.
Trimming the intervals encourages the iterates to stay in the interior of the domain.
Second, we reject $\xv_{t+1}$ if it violates the constraints and stay at $\xv_t$.
In the next iteration, a new ellipse is sampled (with a different $\nuv$) and hopefully the new ellipse has a nicer intersection with the domain that is less prone to numerical issues.
The rejection occurs rarely and is only used to safeguard against these numerical issues.
Therefore, this ``rejection'' is different from the ``rejection'' in rejection sampling.
Indeed, the ``rejection'' rate is practically zero in the experiments.
Combining those two tricks leads to a robust implementation in single precision floating-point arithmetic.

Another way to enhance the numerical stability is simply using a higher floating-point precision, \eg, double precision.
Empirically, we find elliptical slice sampling numerically robust in double precision and no constraint violation has ever occurred in any experiments so far.
However, double precision increases the wall-clock running time, and thus is not recommended unless absolute necessary.

%% file: algorithms/best.tex
\begin{algorithm}[t]
\DontPrintSemicolon

\KwIn{$I_i = \sbb{0, \alpha_i} \cup \sbb{\beta_i, 2\pi}$ with $\alpha_i < \beta_i$ for $i = 1, 2, \cdots, m$}
\KwOut{$I_\act = \cap_{i=1}^{m} I_i$}

sort $\cbb{\alpha_i}_{i=1}^{m}$ in ascending order:
$0 \leq \alpha_{i_1} \leq \alpha_{i_2} \leq \cdots \leq \alpha_{i_m} \leq 2\pi$

compute $\gamma_k = \max\cbb{\beta_{i_1}, \beta_{i_2}, \cdots, \beta_{i_k}}$ for $k = 1, 2, \cdots m$
\tcp*{the cumulative max of $\cbb{\beta_{i_k}}_{k=1}^{m}$}

\Return
$[0, \alpha_{i_1}] \cup \bb[\big]{\bigcup_{k=2}^{m} [\gamma_{k-1}, \alpha_{i_k}]} \cup [\gamma_m, 2\pi]$
\tcp*{define $[\gamma_{k-1}, \alpha_{i_k}] = \emptyset$ if $\gamma_{k-1} > \alpha_{i_k}$}

\caption{Constructing the Active Intervals Analytically}
\label{alg:best}
\end{algorithm}

%% file: figures/intersections/ratio_gt_1.tex
\begin{tikzpicture}[scale=0.7]
\begin{axis}[
    x=1cm,y=1cm, grid=major,
    xmin=-3, xmax=3,
    ymin=-2, ymax=2,
]

\def\xtx{2}
\def\xty{0}
\def\nux{0}
\def\nuy{1}

\def\ex#1{{\xtx * cos(deg(#1)) + \nux * sin(deg(#1)}}
\def\ey#1{{\xty * cos(deg(#1)) + \nuy * sin(deg(#1)}}

\coordinate[label=below right:$O$] (O) at (0, 0);
\coordinate (xt) at (\xtx, \xty);
\coordinate (nu) at (\nux, \nuy);

\draw[->,very thick, dashed] (O) -- node[below, pos=0.6] {$\xv_t$} (xt);
\draw[->,very thick, dashed] (O) -- node[left, pos=0.5] {$\nuv_t$} (nu);

\coordinate (A) at (-2.5, -1);
\coordinate (B) at (-2.5, 1);

\draw [name path=L1,very thick, blue] ($(B)!6cm!(A)$) -- ($(A)!6cm!(B)$);

\coordinate (southwest) at (-3, -3);
\coordinate (southeast) at (3, -3);
\coordinate (northeast) at (3, 3);
\coordinate (northwest) at (-3, 3);

\path[name path=bottom] (southwest) -- (southeast);
\path[name path=top] (northwest) -- (northeast);
\path[name path=right] (northeast) -- (southeast);

\path[name intersections={of=L1 and top, by={i1}}];
\path[name intersections={of=L1 and bottom, by={i2}}];

\draw[fill=blue, opacity=0.2] (i1) -- (i2) -- (southeast) -- (northeast);

\addplot[domain=0:2 * pi,samples=100,red,ultra thick] (\ex{x}, \ey{x});
\end{axis}

\end{tikzpicture}

%% file: figures/intersections/ratio_=_1.tex
\begin{tikzpicture}[scale=0.7]
\begin{axis}[
    x=1cm,y=1cm, grid=major,
    xmin=-3, xmax=3,
    ymin=-2, ymax=2,
]

\def\xtx{2}
\def\xty{0}
\def\nux{0}
\def\nuy{1}

\def\ex#1{{\xtx * cos(deg(#1)) + \nux * sin(deg(#1)}}
\def\ey#1{{\xty * cos(deg(#1)) + \nuy * sin(deg(#1)}}

\coordinate[label=below right:$O$] (O) at (0, 0);
\coordinate (xt) at (\xtx, \xty);
\coordinate (nu) at (\nux, \nuy);

\draw[->,very thick, dashed] (O) -- node[below, pos=0.6] {$\xv_t$} (xt);
\draw[->,very thick, dashed] (O) -- node[left, pos=0.5] {$\nuv_t$} (nu);

\coordinate (A) at (-2, -1);
\coordinate (B) at (-2, 1);

\draw [name path=L1,very thick, blue] ($(B)!6cm!(A)$) -- ($(A)!6cm!(B)$);

\coordinate (southwest) at (-3, -3);
\coordinate (southeast) at (3, -3);
\coordinate (northeast) at (3, 3);
\coordinate (northwest) at (-3, 3);

\path[name path=bottom] (southwest) -- (southeast);
\path[name path=top] (northwest) -- (northeast);
\path[name path=right] (northeast) -- (southeast);

\path[name intersections={of=L1 and top, by={i1}}];
\path[name intersections={of=L1 and bottom, by={i2}}];

\draw[fill=blue, opacity=0.2] (i1) -- (i2) -- (southeast) -- (northeast);

\addplot[domain=0:2 * pi,samples=100,red,ultra thick] (\ex{x}, \ey{x});
\draw[fill,red] (\ex{pi}, \ey{pi}) circle (3pt);
\end{axis}

\end{tikzpicture}

%% file: figures/intersections/ratio_lt_1.tex
\begin{tikzpicture}[scale=0.7]
\begin{axis}[
    x=1cm,y=1cm, grid=major,
    xmin=-3, xmax=3,
    ymin=-2, ymax=2,
]

\def\xtx{2}
\def\xty{0}
\def\nux{0}
\def\nuy{1}

\def\ex#1{{\xtx * cos(deg(#1)) + \nux * sin(deg(#1)}}
\def\ey#1{{\xty * cos(deg(#1)) + \nuy * sin(deg(#1)}}

\coordinate[label=below right:$O$] (O) at (0, 0);
\coordinate (xt) at (\xtx, \xty);
\coordinate (nu) at (\nux, \nuy);

\draw[->,very thick, dashed] (O) -- node[below, pos=0.6] {$\xv_t$} (xt);
\draw[->,very thick, dashed] (O) -- node[left, pos=0.5] {$\nuv_t$} (nu);

\coordinate (A) at (\ex{pi * 2 / 3}, \ey{pi * 2 / 3});
\coordinate (B) at (\ex{pi * 2 / 3}, 1);

\draw [name path=L1,very thick, blue] ($(B)!6cm!(A)$) -- ($(A)!6cm!(B)$);

\coordinate (southwest) at (-3, -3);
\coordinate (southeast) at (3, -3);
\coordinate (northeast) at (3, 3);
\coordinate (northwest) at (-3, 3);

\path[name path=bottom] (southwest) -- (southeast);
\path[name path=top] (northwest) -- (northeast);
\path[name path=right] (northeast) -- (southeast);

\path[name intersections={of=L1 and top, by={i1}}];
\path[name intersections={of=L1 and bottom, by={i2}}];

\draw[fill=blue, opacity=0.2] (i1) -- (i2) -- (southeast) -- (northeast);

\addplot[domain=-0.5 * pi - 1 / 6 * pi:2 / 3 * pi,samples=100,red,ultra thick] (\ex{x}, \ey{x});
\addplot[domain=2 / 3 * pi:4 / 3 * pi,samples=100,black,ultra thick] (\ex{x}, \ey{x});
\draw[fill,red] (\ex{-0.5 * pi - 1 / 6 * pi}, \ey{-0.5 * pi - 1 / 6 * pi}) circle (3pt);
\draw[fill,red] (\ex{2 / 3 * pi}, \ey{2 / 3 * pi}) circle (3pt);
\end{axis}

\end{tikzpicture}

%% file: figures/intersections/ratio_=_-1.tex
\begin{tikzpicture}[scale=0.7]
\begin{axis}[
    x=1cm,y=1cm, grid=major,
    xmin=-3, xmax=3,
    ymin=-2, ymax=2,
]

\def\xtx{2}
\def\xty{0}
\def\nux{0}
\def\nuy{1}

\def\ex#1{{\xtx * cos(deg(#1)) + \nux * sin(deg(#1)}}
\def\ey#1{{\xty * cos(deg(#1)) + \nuy * sin(deg(#1)}}

\coordinate[label=below right:$O$] (O) at (0, 0);
\coordinate (xt) at (\xtx, \xty);
\coordinate (nu) at (\nux, \nuy);

\draw[->,very thick, dashed] (O) -- node[below, pos=0.6] {$\xv_t$} (xt);
\draw[->,very thick, dashed] (O) -- node[left, pos=0.5] {$\nuv_t$} (nu);

\coordinate (A) at (2, -1);
\coordinate (B) at (2, 1);

\draw [name path=L1,very thick, blue] ($(B)!6cm!(A)$) -- ($(A)!6cm!(B)$);

\coordinate (southwest) at (-3, -3);
\coordinate (southeast) at (3, -3);
\coordinate (northeast) at (3, 3);
\coordinate (northwest) at (-3, 3);

\path[name path=bottom] (southwest) -- (southeast);
\path[name path=top] (northwest) -- (northeast);
\path[name path=right] (northeast) -- (southeast);

\path[name intersections={of=L1 and top, by={i1}}];
\path[name intersections={of=L1 and bottom, by={i2}}];

\draw[fill=blue, opacity=0.2] (i1) -- (i2) -- (southeast) -- (northeast);

\addplot[domain=0:2 * pi,samples=100,black,very thick] (\ex{x}, \ey{x});
\draw[fill,red] (\ex{0}, \ey{0}) circle (3pt);
\end{axis}

\end{tikzpicture}

%% file: figures/intersections/ratio_lt_-1.tex
\begin{tikzpicture}[scale=0.7]
\begin{axis}[
    x=1cm,y=1cm, grid=major,
    xmin=-3, xmax=3,
    ymin=-2, ymax=2,
]

\def\xtx{2}
\def\xty{0}
\def\nux{0}
\def\nuy{1}

\def\ex#1{{\xtx * cos(deg(#1)) + \nux * sin(deg(#1)}}
\def\ey#1{{\xty * cos(deg(#1)) + \nuy * sin(deg(#1)}}

\coordinate[label=below right:$O$] (O) at (0, 0);
\coordinate (xt) at (\xtx, \xty);
\coordinate (nu) at (\nux, \nuy);

\draw[->,very thick, dashed] (O) -- node[below, pos=0.6] {$\xv_t$} (xt);
\draw[->,very thick, dashed] (O) -- node[left, pos=0.5] {$\nuv_t$} (nu);

\coordinate (A) at (2.5, -1);
\coordinate (B) at (2.5, 1);

\draw [name path=L1,very thick, blue] ($(B)!6cm!(A)$) -- ($(A)!6cm!(B)$);

\coordinate (southwest) at (-3, -3);
\coordinate (southeast) at (3, -3);
\coordinate (northeast) at (3, 3);
\coordinate (northwest) at (-3, 3);

\path[name path=bottom] (southwest) -- (southeast);
\path[name path=top] (northwest) -- (northeast);
\path[name path=right] (northeast) -- (southeast);

\path[name intersections={of=L1 and top, by={i1}}];
\path[name intersections={of=L1 and bottom, by={i2}}];

\draw[fill=blue, opacity=0.2] (i1) -- (i2) -- (southeast) -- (northeast);

\addplot[domain=0:2 * pi,samples=100,black,very thick] (\ex{x}, \ey{x});
\end{axis}

\end{tikzpicture}

%% file: figures/intersections/legend.tex
\begin{tikzpicture}
\matrix [draw,below left] at (0, 0) {
  \draw[red,ultra thick] (0, 0) -- (1, 0) node[right,black] {Ellipse-halfspace intersection}; \\
  \draw[blue, very thick] (0, 0) -- (1, 0) node[right, black] {Hyperplane $\av^\top \xv = b$}; \\
  \fill[blue, opacity=0.2] (0, 0) rectangle (1, -0.55) node[above right,black,opacity=1] {Halfspace $\cbb{\xv \in \Rb^d: \av^\top \xv \leq \bv}$}; \\
};
\end{tikzpicture}

%% file: paper/experiments.tex
All simulations are run on a single machine with a RTX 3090 GPU using single precision floating points.
We use BoTorch \textsc{v0.11.1} as the baseline.
Our code is available at \url{https://github.com/kayween/linear-ess}.

\subsection{One Dimensional Truncated Normal Sampling}
We run elliptical slice sampling with \Cref{alg:best} on two univariate truncated normal distributions:
$\Nc(0, 1)$ truncated by $-1 \leq x \leq 3$ and $15 \leq x \leq 16$, respectively.
It is well-known that the density and moments of a univariate truncated normal can be computed in closed-forms, which will serve as sanity checks.

We draw $10^5$ samples from each distribution by running $2000$ independent Markov chains in parallel.
We use $500$ iterations of burn-in and a thinning of $10$.
Hence, the total number of steps is $2000 \times 500 + 10 \times 10^5 = 2 \times 10^6$.
The mean and variance estimates are accurate at least to the second digit after the decimal point.
For the truncated normal $\Nc(0, 1) \in [-1, 3]$, no rejection happens.
On the other hand, a total of $8$ rejections occurs for the truncated normal $\Nc(0, 1) \in [15, 16]$ out of $2 \times 10^6$ Markov chain steps ($\approx 0.0004\%$ rejection rate).
Note that the density of the truncated normal $ \Nc(0, 1) \in [15, 16]$ concentrates around $15$.
As a result, the Markov chain is forced to stay close to the domain boundary with high probability, which imposes a serious numerical challenge to elliptical slice sampling.
Indeed, the rejection discussed in \S\ref{sec:numerical-stability} is necessary to safeguard the algorithm in extreme situations.

\begin{figure}[t]
\centering
\begin{subfigure}[b]{0.3\linewidth}
\vspace{0pt}
\includegraphics[width=\linewidth]{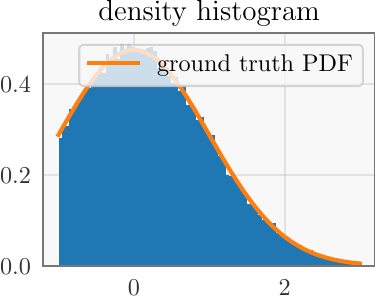}
\caption{$-1 \leq \Nc(0, 1) \leq 3$}
\label{fig:univariate-mcmc-first}
\end{subfigure}
\begin{subfigure}[b]{0.3\linewidth}
\vspace{0pt}
\includegraphics[width=\linewidth]{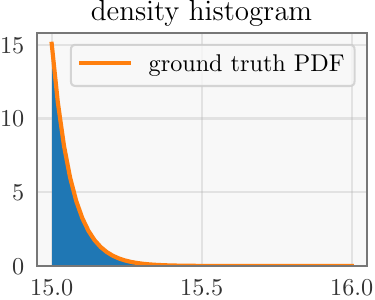}
\caption{$15 \leq \Nc(0, 1) \leq 16$}
\label{fig:univariate-mcmc-second}
\end{subfigure}
\begin{subfigure}[b]{0.36\linewidth}
\centering
\begin{tabular}{ccc}
\toprule
truncation & $[-1, 3]$ & $[15, 16]$ \\
\midrule
true $\mu$ & $0.2828$ & $15.0661$ \\
estimates $\hat\mu$ & $0.2820$ & $15.0661$ \\
\cmidrule{1-3}
true $\sigma^2$ & $0.6161$ & $0.0043$ \\
estimates $\hat\sigma^2$ & $0.6148$ & $0.0043$ \\
\cmidrule{1-3}
``rejections'' & 0 & 8 \\
\bottomrule
\end{tabular}
\vspace{1.2em}
\caption{Estimated statistics.}
\end{subfigure}
\caption{
Draw $10^5$ samples from univariate truncated normal distributions with parallel Markov chains.
}
\end{figure}

\subsection{Accelerate High Dimensional Truncated Normal Sampling}

We demonstrate \Cref{alg:best} accelerates high dimensional truncated normal sampling, especially when the number of inequality constraints $m$ is large.
We generate a set of random instances with varying dimensions as follows.
First, we generate a $d \times d$ random matrix $\Av$ whose entries are \iid samples from a univariate standard normal distribution.
Second, we generate a random vector $\xv_0$ drawn from a $d$-dimensional standard normal distribution, which will used as the initialization the Markov chain.
Third, we set $\bv = \Av \xv_0 + \uv$, where $\uv$ is a random vector drawn uniformly from the hypercube $[0, 1]^d$.
By construction, the initialization $\xv_0$ lies in the interior of the domain.
Note that the number of constraints $m = d$ increases as the number of dimensions increases.

\begin{figure}[t]
\centering
\begin{subfigure}{0.33\linewidth}
\includegraphics[width=\linewidth]{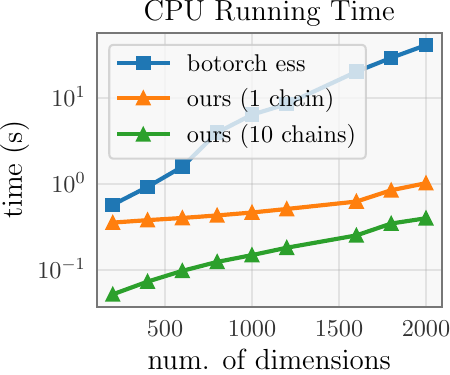}
\end{subfigure}
\begin{subfigure}{0.33\linewidth}
\includegraphics[width=\linewidth]{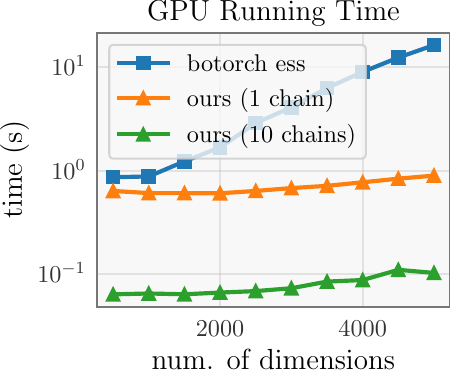}
\end{subfigure}
\caption{Running time of drawing $1000$ samples from high dimensional truncated normal distributions.}
\label{fig:high-dimensional-truncated}
\end{figure}

In \Cref{fig:high-dimensional-truncated}, we draw $1000$ samples from the general instances of truncated normal distributions and compare the running time against BoTorch's implementation.
BoTorch's implementation runs a single Markov chain for $1000$ steps.
Our implementation runs either a single Markov chain for $1000$ steps or $10$ chains in parallel for $100$ steps.
Both of them use no burn-in and no thinning.
No rejection occurs when running our implementation on these high dimensional distributions.
With a single Markov chain, our implementation is over $10$x faster than BoTorch's implementation in high dimensions, \eg, $d \geq 1000$ on CPU and $d \geq 4000$ on GPU.
This speed-up solely comes from the improved per iteration complexity $\Oc(m \log m)$.
Furthermore, running $10$ Markov chains on GPU in parallel yields an additional $10$x speed-up in high dimensions.

%% file: paper/related.tex
Truncated normal sampling is hard as soon as the number of dimensions $d \geq 2$, and thus has to rely on Monte Carlo methods.
Elliptical slice sampling was originally proposed by \textcite{murray2010elliptical} as a general slice sampling method with normal priors.
The rejection-free version discussed in this paper is an adaptation based on \textcite{fagan2016elliptical,gessner2020integrals}.
Other Markov chain Monte Carlo methods, when adapted to linearly truncated normal distributions, often have efficient rejection-free updates as well, \eg, Gibbs sampling \parencite{kroese2013handbook}, Hamiltonian Monte Carlo \parencite{pakman2014exact}.

Related to truncated sampling is numerically estimating the integral $Z = \int_{\xv \in \Dc} \phi(\xv) \diff \xv$, \ie, the normal probability of the domain $\Dc$.
When the domain $\Dc$ is axis-aligned, this integral is exactly the cumulative distribution function of multivariate normal distributions, which unfortunately has no closed-form expression unless the coordinates are independent.
Many numerical methods estimating the normal probability are based on separation of variables \parencite{genz1992numerical}, including recent developments like bivariate conditioning \parencite{genz2016numerical} and minimax tilting \parencite{botev2017normal}.
However, methods based on separation of variables often do not work with an arbitrary number of inequality constraints, especially when $m > 2 d$.
A general Monte Carlo estimator, developed recently by \textcite{gessner2020integrals}, is based on the Holmes-Diaconis-Ross algorithm \parencite{kroese2013handbook} and elliptical slice sampling \parencite{murray2010elliptical}, which has witnessed several machine learning applications in the last few years \parencite[\eg,][]{theisen2021evaluating,theisen2021good}.

%% file: paper/conclusion.tex
We have presented a $\Oc(m \log m)$ algorithm computing the active intervals in linear elliptical slice sampling for linearly truncated normal distributions.
We also have discussed extensively how to handle edge cases.
We hope our algorithm and implementation unlock the full potential of elliptical slice sampling for linearly truncated normal distributions,
and enable new applications that are previously bottlenecked by the speed of sampling.

We end this paper by mentioning two extensions.
First, it is interesting to adapt elliptical slice sampling to handle linear equality constraints, in which case the Markov chain has to run in the null space of the linear equality constraints.
Second, it is interesting to support differentiable samples by adapting the idea of \textcite{zoltowski2021slice}.

%% file: appendix.tex
\section{Non-Standard Normal Distributions}
\label{sec:nonstandard-normal}
The standard normal assumption is without loss of generality, since non-standard normal distributions reduce to the standard one by a change of variable.
Let $\xv \sim \Nc(\muv, \Sigmav)$ and let $\Lv \Lv^\top = \Sigmav$ be the Cholesky decomposition.
Let $\uv \sim \Nc(\zero, \Iv)$ be a standard normal variable.
Truncating $\xv$ by $\Dc = \cbb{\xv \in \Rb^d: \Av \xv \leq \bv}$ is the same as truncating $\uv$ by a transformed domain $\Dc^\prime = \cbb{\uv \in \Rb^d: \Av \Lv \uv \leq \bv - \Av \muv}$.
Thus, we can sample from the truncated standard normal, truncated by $\Dc^\prime$, and then apply a linear transformation $\uv \mapsto \Lv \uv + \muv$.

\section{Roots of the Trigonometry Equation}
\label{sec:trigonometry}
\input{appendix/trigonometry}

\section{Proofs}
\label{sec:proofs}
\input{appendix/proofs}

\section{Brute Force Intersection Time Complexity}
\label{sec:naive-running-time}
\input{appendix/naive-running-time}

%% file: appendix/trigonometry.tex
This section solves the trigonometry equation
\[
    \av^\top \xv \cos\theta + \av^\top \nuv \sin\theta = b.
\]
Define $p = \av^\top \xv$, $q = \av^\top \nuv$, and $r = \sqrt{p^2 + q^2}$.
WLOG, we assume $r \neq 0$, otherwise the corresponding inequality constraint is either invalid ($b < 0$) or a tautology ($b \geq 0$).
Note that the ratio $b / r \geq -1$, otherwise we have $b < -r \leq \av^\top \xv$.
This causes a contradiction since $\xv \in \Dc$ is a feasible point satisfying the linear inequality $\av^\top \xv \leq \bv$.

Dividing both sides by $r$ gives
\[
    \frac{p}{r} \cos\theta + \frac{q}{r} \sin\theta = \frac{b}{r}.
\]
There exists a unique angle $\tau \in [-\pi, \pi]$ (ignoring the repetition at the boundary) such that $\cos\tau = \frac{p}{r}$ and $\sin\tau = \frac{q}{r}$.
In practice, $\tau$ is given by \verb|arctan2(q, p)|, a function implemented in many libraries.
Applying the angle sum formula gives
\[
    \cos(\theta - \tau) = \frac{b}{r}.
\]
It is clear that the ratio $b / r$ determines the number of roots.
When $-1 < b / r < 1$, the two distinct roots are given by
\begin{equation}
\label{eq:roots}
    \theta = \tau \pm \arccos\bb[\bigg]{\frac{b}{r}},
\end{equation}
A multiple of $2 \pi$ has to be added to the roots, if necessary, to make sure the angles fall in into $[0, 2\pi]$.

Note that \eqref{eq:roots} is not the only form of the roots.
For instance, \textcite{gessner2020integrals} used the roots of the form
\[
    \theta = \pm \arccos\bb[\bigg]{\frac{b}{r}} + 2 \arctan\bb[\bigg]{\frac{q}{r + p}}.
\]
Another root formula, used by \textcite{benavoli2021unified}, is of the form
\[
    \tan\frac12\theta = \frac{q \pm \sqrt{r^2 - b^2}}{b + p}.
\]
Proving these root formulas is left as an exercise for the readers.
Despite their equivalence, we recommend using our root formula \eqref{eq:roots}.
This is an unbiased opinion, since we arrive at this conclusion after trying all formulas.
The other two root formulas need to check additional edge cases when $r + p \approx 0$ and $b + p \approx 0$, which do happen annoyingly in certain extreme situation in practice.

%% file: appendix/proofs.tex
\ClosedFormIntersection*
\begin{proof}
The sorted angles $\cbb{\alpha_{i_k}}_{k=1}^{m}$ divide $[0, 2\pi]$ into $m + 1$ disjoint segments:
\begin{align*}
\sbb{0, 2\pi} =
[0, \alpha_{i_1}] \cup
(\alpha_{i_1}, \alpha_{i_2}] \cup
\cdots \cup
(\alpha_{i_{m-1}}, \alpha_{i_m}] \cup
(\alpha_{i_m}, 2\pi].
\end{align*}
The active intervals $I_\act$ are constructed by computing the intersection of $I_\act$ with each segment.
That is, we use the trivial identity
\begin{align*}
I_\act
= I_\act \cap \sbb{0, 2\pi}
=
\underbrace{
    \bb[\big]{I_\act \cap \sbb{0, \alpha_{i_1}}}
}_{\text{part one}}
\cup
\underbrace{
    \bb*{\bigcup_{k=1}^{m-1} \bb[\big]{I_\act \cap (\alpha_{i_k}, \alpha_{i_{k + 1}}]}}
}_{\text{part two}}
\cup
\underbrace{
    \bb[\big]{I_\act \cap \sbb{\alpha_{i_m}, 2\pi}}
}_{\text{part three}}
\end{align*}
and compute each part analytically.

\textbf{Part One.}
The intersection with the first segment is easy to compute:
\begin{align*}
I_\act \cap \sbb{0, \alpha_{i_1}}
=
\bb*{\bigcap_{i=1}^{m} I_i} \cap \sbb{0, \alpha_{i_1}}
=
\bigcap_{i=1}^{m} \bb[\big]{I_i \cap \sbb{0, \alpha_{i_1}}}
=
\sbb{0, \alpha_{i_1}},
\end{align*}
where the first equality uses the definition of the active intervals $I_\act$;
the second equality swaps the order of intersections; the third equality uses this observation: $\sbb{0, a_{i_1}}$ is a subset of all $I_i$ for $i \in \sbb{m}$ since $\alpha_{i_1}$ is the smallest angle among all $\alpha_i$'s and $\beta_i$'s.

\textbf{Part Three.}
Similarly, the intersection with the last segment is also easy to compute:
\begin{align*}
I_\act \cap (a_{i_m}, 2\pi]
=
\bigcap_{i=1}^{m} \bb[\Big]{
    I_i \cap (\alpha_{i_m}, 2\pi]
}
=
\bigcap_{i=1}^{m} \bb[\Big]{
    \sbb{\beta_i, 2\pi} \cap (\alpha_{i_m}, 2\pi]
}
=
\sbb[\bigg]{\max_{1 \leq i \leq m} \beta_i, 2\pi},
\end{align*}
where the first equality plugs in the definition $I_\act = \cap_{i=1}^{m} I_i$;
the second equality is because $\alpha_{i_m}$ is the largest angle among all $\alpha_{i}$'s and therefore we can ignore $\sbb{0, \alpha_i}$;
the last equality is due to
\[
    \alpha_{i_m} < \beta_{i_m} \leq \max_{1 \leq i \leq m} \beta_i
\]
and thus the chunk $[\alpha_{i_m}, \max_{1 \leq i \leq m} \beta_{i})$ is removed from $\sbb{a_{i_m}, 2\pi}$.

\textbf{Part Two.}
Now we deal with the remaining segments $(a_{i_{k-1}}, a_{i_k}]$ for $k = 2, 3, \cdots, m$.
We assume $a_{i_{k-1}}$ is strictly smaller than $a_{i_k}$ for now and defer the case $a_{i_{k-1}} = a_{i_k}$ to the end.
For a fixed $k$, we must compute
\begin{align*}
I_\act \cap (a_{i_{k-1}}, a_{i_k}]
=
\bb[\Bigg]{\bigcap_{i=1}^{m} I_i} \cap (a_{i_{k-1}}, a_{i_k}]
=
\bigcap_{i=1}^{m} \bb*{I_i \cap (a_{i_{k-1}}, a_{i_k}]}.
\end{align*}
Now we split the intersection index $i$ into two cases $\cbb{i \in [m]: \alpha_i \geq \alpha_{i_k}}$ and $\cbb{i \in [m]: \alpha_i \leq \alpha_{i_{k-1}}}$.
For the first case, notice that
\begin{align*}
    \bigcap_{\cbb{i \in [m]: \alpha_i \geq \alpha_{i_k}}} \bb*{I_i \cap (a_{i_{k-1}}, a_{i_k}]} = (\alpha_{i_{k-1}}, \alpha_{i_k}],
\end{align*}
because the choice of the index $i$ implies $(\alpha_{i_{k-1}}, \alpha_{i_k}] \subseteq \sbb{0, \alpha_i} \subseteq I_i$.
For the second case, we have
\begin{align*}
\bigcap_{\cbb{i \in [m]: \alpha_i \leq \alpha_{i_{k-1}}}} \bb*{I_i \cap (a_{i_{k-1}}, a_{i_k}]}
& =
\bigcap_{\cbb{i \in [m]: \alpha_i \leq \alpha_{i_{k-1}}}} \bb*{\bb[\big]{[0, \alpha_i] \cup [\beta_i, 2\pi]} \cap (\alpha_{i_{k-1}}, \alpha_{i_k}]}
\\
& =
\bigcap_{\cbb{i \in [m]: \alpha_i \leq \alpha_{i_{k-1}}}} \bb*{[\beta_i, 2\pi] \cap (\alpha_{i_{k-1}}, \alpha_{i_k}]}
\\
& =
\sbb*{\max \cbb{\beta_{i_1}, \beta_{i_2}, \cdots, \beta_{i_{k-1}}}, 2\pi} \cap (\alpha_{i_{k-1}}, \alpha_{i_k}],
\end{align*}
where the first equality plugs in the definition of $I_i$;
the second equality is because the index $i$ is specifically chosen such that $\alpha_i \leq \alpha_{i_{k-1}}$;
the last equality is because $\alpha_i$'s are sorted: the indices such that $\alpha_i \leq \alpha_{i_{k-1}}$ are precisely $i_1, i_2, \cdots, i_{k-1}$.

Define $\gamma_k = \max \cbb{\beta_{i_1}, \beta_{i_2}, \cdots, \beta_{i_{k-1}}}$.
Combining the two cases, we obtain
\begin{align*}
I_\act \cap (a_{i_{k-1}}, a_{i_k}] =
\begin{cases}
\sbb{\gamma_k, \alpha_{i_k}} & \text{if} \ \gamma_k \leq \alpha_{i_k}
\\
\emptyset & \text{otherwise}
\end{cases}
\end{align*}
for $k = 2, 3, \cdots, m$.
Finally, we come back to the edge case $a_{i_{k-1}} = a_{i_k}$, which implies $(a_{i_{k-1}}, a_{i_k}] = \emptyset$ by convention.
Thus, the intersection $I_\act \cap (a_{i_{k-1}}, a_{i_k}]$ is automatically empty.
One can verify that the above expression outputs an empty set as well, since $\gamma_k \geq \beta_{i_{k-1}} > \alpha_{i_{k-1}} = \alpha_{i_k}$.
\end{proof}

%% file: appendix/naive-running-time.tex
This section shows that the time complexity of \Cref{alg:naive} is at least $\Omega(m^2)$.
To do so, we construct a worst-case input on which \Cref{alg:naive} takes at least $\Omega(m^2)$ operations.

Without loss of generality, we will work with intervals of the form $I_i = [0, \alpha_i] \cup [\beta_i, 1]$.
Consider the following intersection intervals
\begin{align*}
I_i =
\sbb*{0, \bb*{\frac13}^i} \cup
\sbb*{2 \bb*{\frac13}^i, 1},
\quad i = 1, 2, \cdots, m.
\end{align*}

By induction, it is easy to show that the intersection of the first $k$ intervals is
\begin{align*}
\bigcap_{i=1}^{k} I_i
=
\sbb*{0, \bb[\Big]{\frac{1}{3}}^k} \cup
\bb*{\bigcup_{i=1}^{k} \sbb*{2 \bb*{\frac13}^i, \bb*{\frac13}^{i-1}}}.
\end{align*}
In particular, $\cap_{i=1}^{k} I_i$ is an union of $k + 1$ intervals.
Thus, the $k$-th inner loop of \Cref{alg:naive} takes $\Omega(k)$ operations.
As a result, the algorithm takes $\Omega(m^2)$ operations in total.
It is also not hard to verify that presorting $\cbb{\alpha_i}_{i=1}^{m}$, or $\cbb{\beta_i}_{i=1}^{m}$, does not reduce the time complexity.
The worse-case inputs for these variants can be constructed similarly.